# On Leakage in Machine Learning Pipelines

Sasse, L.*[1,2,3]; Nicolaisen-Sobesky, E.*[1,2]; Dukart, J.[1,2]; Eickhoff[1,2], S.B.; Götz, M.[4,5]; Hamdan, S.[1,2]; Komeyer, V.[1,2,6]; Kulkarni, A.[7]; Lahnakoski, J.[1,2]; Love, B.C.[8,9,10]; Raimondo, F.[1,2]; Patil, K.R.[1,2,11]

1. Institute of Neuroscience and Medicine, Brain and Behaviour (INM-7), Forschungszentrum Jülich, Jülich, Germany
2. Institute of Systems Neuroscience, Medical Faculty, Heinrich-Heine-University Düsseldorf, Düsseldorf, Germany
3. Max Planck School of Cognition, Stephanstrasse 1a, Leipzig, Germany
4. Division of Experimental Radiology, Department for Diagnostic and Interventional Radiology, University Hospital Ulm, Ulm, Germany
5. Experimental Radiology, University Ulm, Ulm, Germany
6. Department of Biology, Faculty of Mathematics and Natural Sciences, Heinrich Heine University Duesseldorf, Duesseldorf, Germany
7. Principal Global Services, Pune, India
8. Department of Experimental Psychology, University College London, London, UK
9. The Alan Turing Institute, London, UK
10. European Lab for Learning & Intelligent Systems (ELLIS)
11. Koita Centre for Digital Health, IIT Bombay, Mumbai 400076, India

* These authors contributed equally
Corresponding author: Dr. Kaustubh R. Patil (k.patil@fz-juelich.de)

**Abstract:**

Machine learning (ML) provides powerful tools for predictive modeling. ML's popularity stems from the promise of sample-level prediction with applications across a variety of fields from physics and marketing to healthcare. However, if not properly implemented and evaluated, ML pipelines may contain leakage typically resulting in overoptimistic performance estimates and failure to generalize to new data. This can have severe negative financial and societal implications. Our aim is to expand understanding associated with causes leading to leakage when designing, implementing, and evaluating ML pipelines. Illustrated by concrete examples, we provide a comprehensive overview and discussion of various types of leakage that may arise in ML pipelines.

# 1. Introduction

Machine learning (ML) has become a popular approach to make predictions, aid decision making, and gain insights into complex data. Various methodologies like supervised, unsupervised, generative, and reinforcement learning define the ML landscape, each with its own unique strengths and applications. In supervised learning, the machine learns a function that links input to output by utilizing labeled training data where the correct output is known, derived from example input-output pairs [34]. Unsupervised learning deals with unlabelled data. The machine must figure out the correct answer without being told about a ground truth and must therefore discover patterns and structures in the input data (e.g. using clustering) [34]. Generative learning is a machine learning approach centered on generating novel data samples. This technique is commonly applied in tasks like producing images, text, and various other data types [23]. Reinforcement learning involves an agent interacting with an environment, taking actions, and receiving rewards or penalties. Through repeated interactions, the model autonomously learns the optimal strategy to maximize rewards, relying less on external guidance for output determination [67].

However, despite the due use and applicability of other methods, supervised learning reigns as the most crucial for prediction and practical use, due to its capacity to build predictive models with applications in various domains including health-care, physics, and climate science [6, 10, 15, 43, 57, 66, 68, 74, 80]. This is not only because supervised learning is well suited to the tabular data that most scientific problem domains are likely to collect, but also because easy-to-use software libraries with hundreds of learning algorithms and data wrangling tools have lowered the entry barrier for ML-based analyses (e.g. scikit-learn [54] and tidymodels [42]). These collaborative advancements in accessible tools, expanding datasets, and evolving methodologies demonstrate the considerable promise of ML applications to drive transformative innovation across diverse problem domains.

Despite the availability of easy-to-use ML software, most applications still require assembling a custom ML-based data analysis pipeline satisfying unique considerations in terms of data, preprocessing, feature engineering, parameter tuning, and model selection. While end-to-end tools exist, opting for these often sacrifices control for convenience (e.g. [11]). Therefore, implementing a correct ML pipeline and drawing valid conclusions from the ensuing results remains challenging, and prone to errors. This challenge extends beyond technical aspects, impacting the interpretability and trustworthiness of the outcomes. Handcrafted pipelines, although demanding, afford practitioners the precision, control and insight required for complex data analysis scenarios. Additionally, the evolving nature of data and algorithmic advancements continually reshapes best practices, necessitating a balance between automation and custom solutions to ensure accuracy and relevance in analyses. Striking this balance remains pivotal for robust, reliable, and impactful ML applications.

ML models are powerful, and they are adept at exploiting any available information. Thus, it falls on the practitioner to ensure that the modeling approach is reliable and valid. As we will discuss, even simple ML pipelines, if not properly implemented and interpreted, can lead to drastically wrong interpretations and severely problematic conclusions. These issues extend far beyond academic debates; they hold immense societal relevance. Widespread adoption of flawed practices in machine learning can exact substantial societal and economic costs, underscoring the urgency to rectify and mitigate these risks [37]. Similar to the replication crisis that recently engulfed the statistics communities and much of the applied sciences, owing to misunderstanding and –intentional or unintentional– misuse of *p*-values from null hypothesis significance testing [32, 78], misunderstandings and malpractice in ML can lead to its own replication crisis [25, 36] with severe negative financial and societal ramifications [76]. It must be noted that reproducibility of a ML pipeline is not sufficient to resolve this as a reproducible ML pipeline could be still incorrect in inference. Addressing such ML pitfalls, and especially pitfalls related to data leakage, is essential to improve the quality and trustworthiness of ML-based data analyses, and consequently to better applications and fostering societal acceptance.

One of the most common and most critical types of error when applying ML is data leakage. Data leakage refers to the leakage of "illegitimate" information into the training process of a ML model [38]. For example, leakage occurs when the model gets to learn from information about the supposed test set and therefore a fair evaluation of the generalization error is not possible, as the test set does not really represent new, unseen data anymore. This likely means that any estimate of the error will be overly optimistic. For instance, a recent study claimed high accuracy (91%) in predicting suicidality in youth using neuroimaging data [35]. Such a model would be of high clinical relevance and can provide valuable insights about underlying brain phenotypes. However, this paper was retracted because it relied on leakage-prone feature selection leading to an erroneous and overfitted model and interpretations [16, 76]. This exemplifies the threat posed by leakage in ML pipelines to realistic estimation of generalization performance, insights gained, and deployment. Although some types of leakage are widely recognized and are discussed in the literature, such as illegitimate use of the targets of the test data [38], many remain unexplored. Leakage can happen in numerous ways; some are evident and straightforward to detect, whereas others can be more subtle, complicating their detection and rectification.

While previous works have addressed several pitfalls in ML-based analysis [5, 45, 59, 61, 70, 73, 74, 79], only a few have covered the wide range of threats posed by leakage [36, 38] (also see John Langford: https://hunch.net/?p=22). We expand these previous works by providing a comprehensive overview of various leakage scenarios, categorized in a user-centric and intuitive fashion which we hope encourages more careful design and evaluation of ML pipelines and inspires further investigation in this area. We aim to equip readers with the necessary tools to effectively recognise leakage in their own (and others') work. This understanding will aid in avoiding these pitfalls, fostering more robust and reliable ML-based analyses.We would like to note that this work does not aim to cover the entire field of machine learning, as it is too vast. For instance, we do not address time series analysis and unsupervised learning, which come with their own unique set of limitations and considerations. However, most of the concepts and guidelines presented here are generally applicable. The authors have noted the misconceptions and malpractices discussed here in open-source code available on the Internet, as well as in code written by themselves, students, or collaborators. These observations span various skill levels, ranging from beginners to domain experts and data analysis experts. Therefore, the insights shared here can provide guidance for everyone from novice to advanced ML practitioners, researchers, reviewers, and editors.

We start with a brief introduction of ML basics and the cross validation (CV) procedure (section 2) that will serve as a guide to understand the concepts used in the rest of the article. This section is divided in three parts: 2.a) ML concepts, 2.b) Cross-validation basics, and 2.c) Steps while designing a ML pipeline. Next, we present various examples of leakage in ML pipelines together with empirical examples and illustrations (section 3). Next, we discuss possible mitigations strategies (section 4) followed by general conclusions and key takeaways (section 5).

## 2. Basics of Machine Learning

2a. Machine Learning concepts

In a supervised machine learning task the user has access to labeled data consisting of $n$ feature-target pairs $S = \{(x_1, y_1), (x_2, y_2), (x_3, y_3), ... (x_n, y_n)\}$ where $x_i$ are the features and $y_i$ are associated targets. The data samples are assumed to be independently and identically distributed (I.I.D.) and sampled from a fixed probability distribution. The task of a ML algorithm is to learn a function or a model that maps features to a target; $f(x_i) = y_i$. A model with discrete output is called a classifier, while one with continuous output is a regressor, two commonly encountered scenarios. The goal is to learn a model that generalizes on unseen data by providing accurate predictions. A model is composed of parameters (e.g., weights in a multiple linear regression) and often includes hyperparameters (e.g., regularization parameter λ of ridge regression). Both contribute significantly to a model's ability to generalize. While the parameters are learned from the data using an optimization procedure, typically involving empirical risk minimization, the hyperparameters need to be either set by the user or "tuned" by searching for values that yield accurate predictions on hold-out data.

*2.b Cross validation basics: model assessment and model selection*

The goal of ML is to create models that accurately predict outcomes on unseen data, which requires learning generalizable information. However, because real-world test data (e.g., future patients or scenarios not yet encountered by a self-driving car) are typically not available, ML practitioners often hold out a portion of the available data as a proxy for test data to evaluate a model's generalization performance. Assuming that the underlying probability distribution of the data does not change, such an estimate helps with *model assessment* as an indicator of what to expect on new data.

Cross validation (CV) is frequently employed for model assessment (Fig. 1) as it makes efficient use of available data [1, 4, 21, 39]. In a $k$-fold CV scheme, the data is divided into $k$ non-overlapping equally sized sets or folds. In each iteration of the CV procedure, one of the folds is used as the test data, while the rest are used for training. Iterating through all folds completes one CV run (also called a repeat). The average performance across all folds is computed as an estimate of generalization performance. To minimize biases that could arise due to data splitting, it is a standard practice to repeat the CV process multiple times with different splits (e.g. 5 times repeated 5-fold CV) [41].

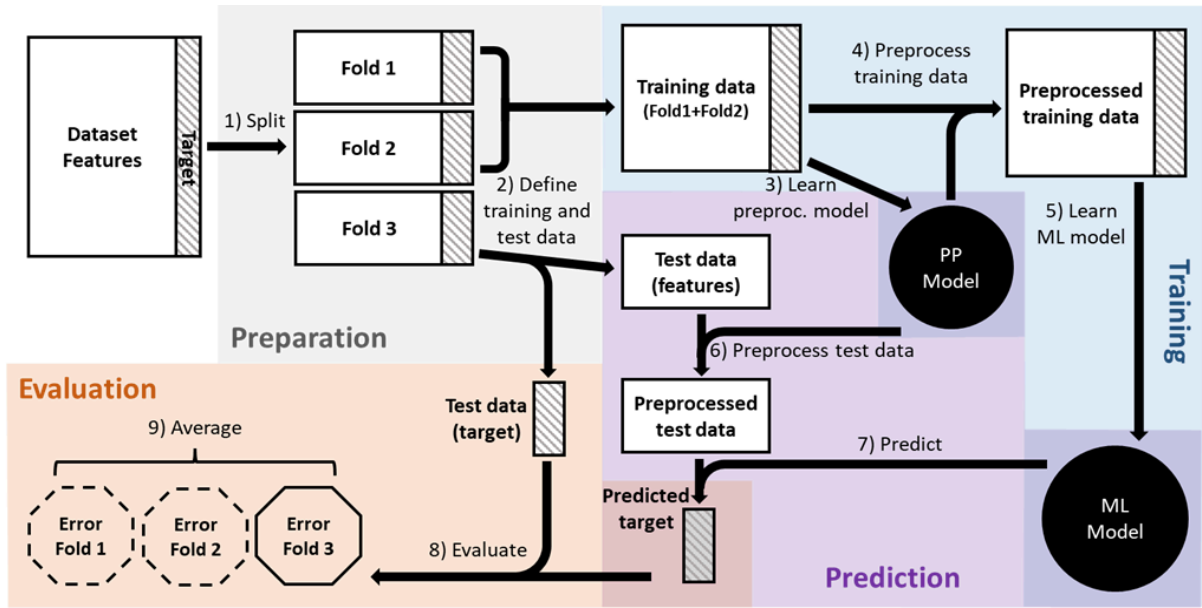

**Figure 1**: A schematic representation of the cross validation (CV) scheme. Here, we illustrate a single repeat of a $k$-fold CV with three folds ($k = 3$) with the third fold being used as the test data.

CV is also employed for *model selection* to select a model from a set of competing options, e.g. models arising from hyperparameter tuning [28] such as the cost of an SVM [83] or pipelines employing different preprocessing and/or learning algorithms. The model with highest generalization performance is typically selected. CV provides a general and practical method for model selection even with complex model parametrization often found in ML algorithms where model selection statistics like Akaike's information criterion might not be feasible. For a more comprehensive coverage of the topic, we refer the reader to excellent sources (e.g. [7, 28, 33]). Both model assessment and model selection are often a part of a ML-based data analysis pipeline. However, as will be discussed in more detail below (see Section 3.a), problems arise if the two roles are confused [41]. Therefore, to cleanly and explicitly differentiate between these two roles of CV (model selection and estimating generalization error), it is necessary to use nested cross-validation [75], also known as double cross-validation. Within a nested CV scheme, the inner CV encompasses all data-dependent decisions and performs model selection (e.g., determining optimal hyperparameters or feature selection) while the outer CV is responsible for model assessment, (i.e. evaluating the model after a finalized model selection on previously completely unseen new data). The key point here is that any decision made on data (i.e. the decision to select a specific model) requires yet again more data, that was not involved in making the decision, to correctly estimate the generalization error.

*2.c Designing a ML pipeline*

The process of designing a ML-based data analysis pipeline can be broadly categorized into the following steps: S-I) Task definition, S-II) Data collection and preparation, S-III) Data preprocessing, S-IV) ML algorithm definition, and S-V) Definition of evaluation scheme and metrics. If the goal of the analysis extends beyond assessing generalization performance, additional steps might be employed, S-VI) Interpretation and deployment. While each of these steps requires multiple decisions that must be made in a data-driven fashion, it is possible and indeed necessary to define how each decision should be made a priori. Mistakes in the data-driven decision making process can lead to data leakage. For more elaborate analysis scenarios, we refer the reader to the CRISP-DM standard [47, 81].

*S-I) Task definition*: Definition of the target variable $y$ (e.g., disease status or behavioral scores) and the features to be used (i.e., $x$, e.g., pixel values in images or functional connectivity derived from neuroimaging data). Consideration of any confounds that can obscure the intended feature-target relationship must be taken into account (e.g., age or sex are often considered as confounds in biological and clinical applications).

*S-II) Data collection and data preparation strategies:* Here decisions need to be made both before and after data collection. Before data collection, decisions to deal with known biases should be made (e.g. equal sampling of males and females, or of case and control observations; see [8, 44] for a detailed treatment of this topic). After collection the data might need preparation. This may involve subsampling (such as selection of only females for sex-specific analysis), feature extraction like connectivity from brain imaging data, and feature preparation like normalization of images. Importantly, we define data preparation as processing exclusively applied to a single data point or sample independently of others. Open and already prepared data are often available and are used directly by many practitioners.

S-III) *Data preprocessing strategy*: Optional data preprocessing steps involving transformations applied across multiple samples are defined. These steps are typically applied to the features, and may include feature normalization, feature selection, dimensionality reduction, and treatment of missing values. Note that domain-specific data preparation and (pre)processing is often employed and the reader is requested to refer to appropriate literature for details.

*S-IV) ML algorithm definition:* One or more ML algorithms suitable for the task at hand must be selected, such as classification for predicting disease status, or regression for predicting continuous behavioral scores. That is, practitioners should a priori define a set of candidate models to involve in model selection. If the ML algorithm includes hyperparameters, the practitioner must either set the hyperparameter values or define a search space and search strategy for tuning them using data (in the model selection process).

*S-V) Definition of evaluation scheme and metrics:* An evaluation scheme must be chosen for model assessment, such as train-test split, $k$-fold CV or use of future data. If the pipeline requires hyperparameter tuning, the chosen scheme should take this into account, for instance by using nested CV. Evaluation metrics appropriate for the task must be selected, such as classification accuracy and area under the receiver operating characteristic curve (AUC) for classification or mean absolute error (MAE) and coefficient of determination ($r^2$) for regression.

S-VI) *Interpretation and deployment:* The selected model can be used to gain insights into the structure of the data. Implicitly interpretable models provide parameters that can be used, e.g., weights of a linear SVM, or additional processing during or post model construction might be needed, e.g., feature importance scores. In real-world application scenarios, the selected pipeline is deployed for making predictions on new samples. In this case, the practitioner must define how the new samples will be acquired and processed before making predictions. While deployment is not considered in typical research settings, as we shall see, it serves as a useful concept for avoiding some potential pitfalls.

**3. Leakage in ML pipelines**

Any data-driven choice made within any step of a ML pipeline (see Section 2c), whether concerning preprocessing, learning, or prediction, must be validated using new unseen data. Failure to use unseen data amounts to leakage and usually results in overly optimistic generalization performance estimates on the data at hand but poor performance on unseen data. We take a general view of leakage to cover inappropriate use of data in different parts of a ML pipeline which can lead to erroneous (either optimistic or pessimistic) estimation of generalization performance or results in non-deployable models. Below, we describe several types of leakage assuming that the ML pipeline employs CV for estimating generalization performance.

*3.a Test-to-train leakage*

We begin with a type of leakage which we call test-to-train leakage as in this case information is leaked from the test set into the training process. Several scenarios can lead to test-to-train leakage, such as failure to separate training and test data, failure to consider dependent samples, application of preprocessing before data splitting, and improper model selection.

The most straightforward case happens when the separation between training and test data is not followed [14], i.e. the test samples as used for training (Figure 2). As the model can learn patterns in the test data, it can result in high test accuracy. However, this cannot be considered a correct estimate of generalization performance as the test data was not unseen. That is, when the separation between training and test data is breached, the model risks fitting to the specific patterns present in the test set. For instance, a k-nearest neighbors (KNN) model (with k=1) will simply remember all the training samples it has seen, and therefore will achieve perfect prediction if the model is "tested" on previously seen training samples. Of course, the resulting error estimate can't possibly hold on truly unseen new data, so that this error estimate is overly optimistic. Another example is optimizing a model to predict a particular test set. This has happened previously in a ML competition where multiple evaluations on the test set were performed, leading to disqualification of the team and consequently withdrawal of the associated paper[1].

---

[1] https://dswalter.github.io/machine-learnings-first-cheating-scandal.html

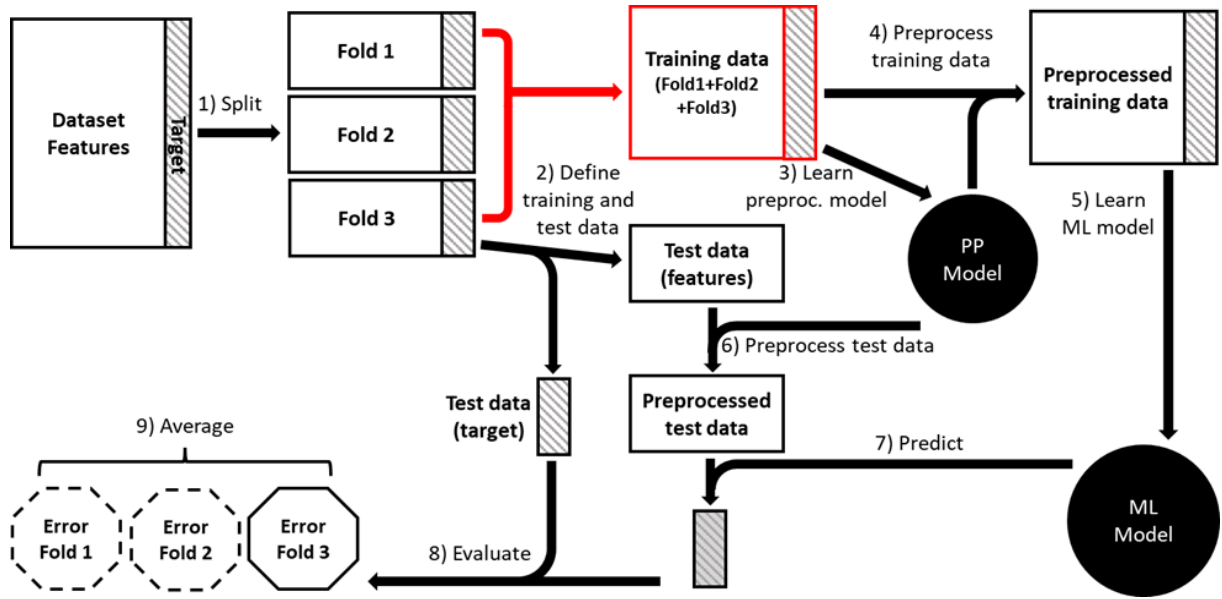

**Figure 2**: Test-to-train leakage: Using the test data as a part of the training data leads to leakage, since the model is able to learn patterns from the test data during training, which usually results in overoptimistic generalization performance estimates. Red color indicates the problematic steps.

Another case where this type of leakage can happen is when the samples are not independent of each other. When data samples can be assumed I.I.D., randomly splitting them into training and test sets is sufficient. However, when the I.I.D. assumption is violated, more care must be taken to avoid related samples being split across training and test sets. An example of violation of the I.I.D. assumption is data that contains twins or siblings as is the case with the Human Connectome Project - Young Adult cohort [72]. Functional Connectivity (FC) as a marker of brain organization is often used as a feature set to predict target variables such as behavioral scores in brain imaging research [19, 29]. Due to heritability, similarity between FC features (and likely also of target variables) is typically higher for twins and siblings than for independently drawn samples [13]. If the twins are allowed to split between training and test sets, this is akin to duplicating the samples albeit noisily, therefore a model can learn about samples in the test set from their siblings in the training set. The prediction performance is therefore higher when splitting family members across folds than if the family members are grouped together (Figure 3). This is an important pitfall, since most often the goal is to build models that will generalize beyond specific families. The ungrouped CV cannot give us an estimate of the error that we would expect for completely new samples, i.e. from new, unseen families. Further examples demonstrating such leakage include, existence of the same genomic loci in the training and test sets when performing cross-cell type predictions [62], and using 2D slices from the 3D brain images of the same individual for training and testing for predicting neurodegenerative disease [82].

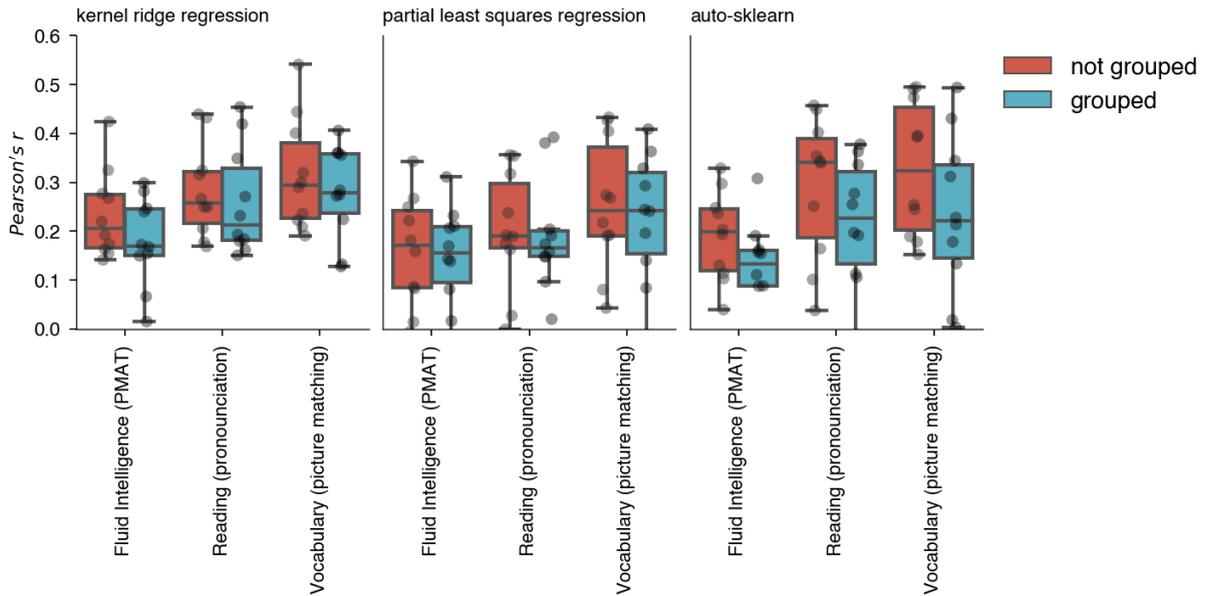

**Figure 3:** Test-to-train leakage due to violation of I.I.D. assumption in cross-validation. Functional connectivity was estimated using resting-state functional magnetic resonance imaging data of the Human Connectome Project - Young Adult cohort (HCP-YA). Functional connectivity was then used to predict three psychometric targets (x-axis), each in a 10-fold cross-validation scheme. HCP-YA data contain siblings, and siblings are known to have similar connectomes. Therefore, allowing the siblings to be split across training and test sets (red bars, without grouping) leads to leakage while grouping siblings in training or test sets (blue bars, with grouping) shows overall lower accuracy (Pearson's r between true and predicted target, y-axis).

In addition to leakage due to the same or similar samples, data leakage can also happen via modeling parameters. A common case of such test-to-train leakage arises when data preprocessing, such as dimensionality reduction (e.g. PCA), feature normalization, and imputation for filling in missing values, is applied to the whole dataset before splitting it for CV [36]. Practitioners may not immediately recognise this as leakage, since the ML model is trained after splitting the data. However, estimating the preprocessing parameters on the whole dataset invalidates the train-test separation (Figure 4). That is, data in the training set is transformed dependent on data in the test set, and crucially, ML models can exploit this to learn about the test set. Therefore, the resulting estimate of generalization performance is likely to be overly optimistic. Empirical demonstrations of such leakage in the literature include performing feature selection on the whole dataset (Figure 5) [60, 70], and oversampling to counter data imbalance [61]. It should be noted that such leakage can happen when preprocessing either the features or the target values. For instance, when the target is created by combining multiple variables (e.g., several behavioral measures) using a dimensionality reduction method such as PCA.

This case of test-to-train leakage can be avoided by learning the preprocessing parameters on the training set and then applying them to the training and the test sets. For example, if one wants to apply feature selection, one should select features based on the training set after splitting data. Crucially, this means that each iteration of CV will involve its own feature selection process which may select a different set of features. Since this implies more computation and also makes interpreting the results more difficult since it is necessary to keep track of different features in different CV iterations, practitioners sometimes erroneously avoid splitting the data before data preprocessing. Note that data preparation strategies that rely on a single sample and thus preserve the train-test separation do not lead to such leakage. For instance, data imputation can be performed in a within-sample fashion such that missing values are estimated using other features of that sample (see, e.g., [18]).

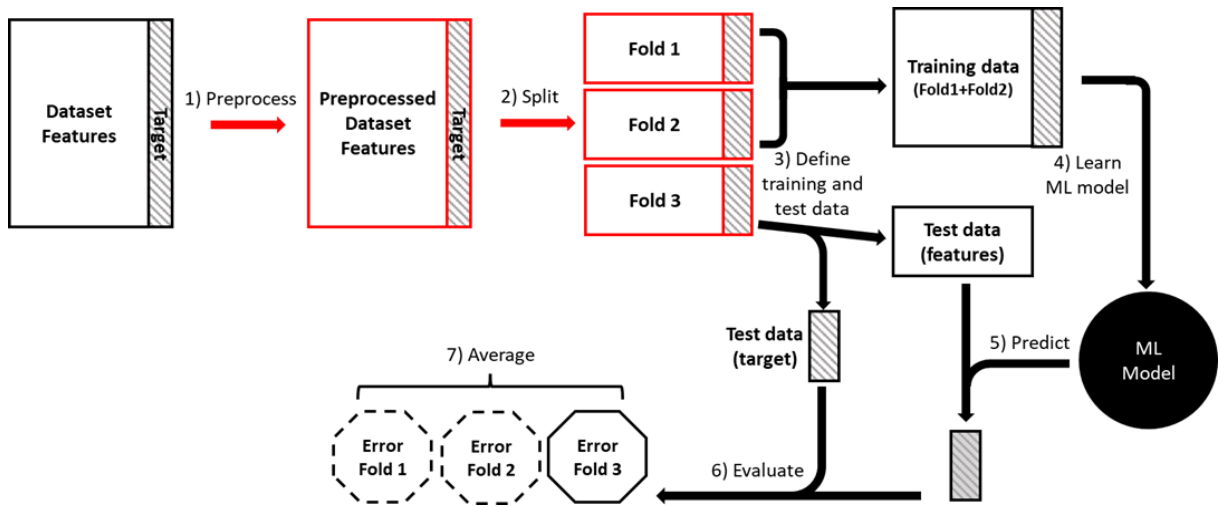

**Figure 4**: Test-to-train leakage. Preprocessing is performed on the whole data before data splitting, yielding a preprocessing model learned from the whole dataset. In a correct implementation, the parameters of a preprocessing model should be estimated in the training set and applied to both training and test sets. Red color indicates the problematic steps.

```
# leakage
features_to_use  = feature_select(X, y)
CV_error = run_cross_validation(X[:, features_to_use], y, k=5)

# no leakage
for k in 1:5
    train_idx = training_set_indices(k)
    test_idx = training_set_indices(k)
    features_to_use = feature_select(X[train_idx, :], y[train_idx])
    model = train_model(X[train_idx, features_to_use], y[train_idx])
    CV_error[k] = test(model, X[test_idx, features_to_use])
```

**Figure 5:** Pseudocode for leakage-inducing and leakage-free feature selection.

A particular case of test-to-train leakage occurs when conflating the two roles of CV, model assessment and model selection (see Section 2.b). Although not immediately obvious, this can be considered test-to-train leakage, because by running CV for many different models, and subsequently making a decision based on the results (i.e. selecting one model among them), the test folds used in the CV essentially become part of the training data. It is important to highlight here that training data does not just refer to data that an algorithm uses to fit parameters but also to data that researchers use to make data-driven decisions. Therefore, the error estimate from the model selection CV is not a valid estimate of true generalization performance. For instance, consider an ML algorithm with a single hyperparameter (e.g., linear kernel SVM with its hyperparameter called cost within a 5-fold CV. For each fold the cost value that provides the lowest error on the set is used. The CV estimate of error is calculated by averaging across the test sets. As a result, the error obtained during model selection is likely an overoptimistic estimate of the generalization error (model assessment) [41, 55, 70, 73]. Using nested CV in which the hyperparameter is tuned in an inner CV and the selected model is applied on the test set avoids such leakage. One of the questions a practitioner may face concerns the fact that different models and hyperparameters are selected in each fold of the CV. Students wonder how they can report and use that model, since there is no such thing as "the model". However, this point of view fails to acknowledge that simply fitting parameters of a model in each iteration of CV will also always result in different models. Instead, nested CV presents an opportunity rather than a challenge, since researchers can then easily test the stability of fitted parameters and hyperparameters chosen in the model selection process over multiple iterations by inspecting each trained (and selected) model. We provide an empirical illustration, again using neuroimaging data to predict behavior, that indeed shows that CV estimates are overoptimistic compared to nested CV estimates (Figure 6).

Such leakage is not restricted to hyperparameter tuning and can happen with any data-driven choices, such as selection of an algorithm (e.g., SVM versus random forests) and data transformations (e.g., PCA, univariate feature selection) [41]. Such choices should be treated in the same way as hyperparameters, i.e. tuned and evaluated using nested CV. In other words, all data-driven choices within a ML pipeline should be considered as a part of learning, hence they must be validated on data not seen by the models to correctly estimate generalization performance. For more details see [71].

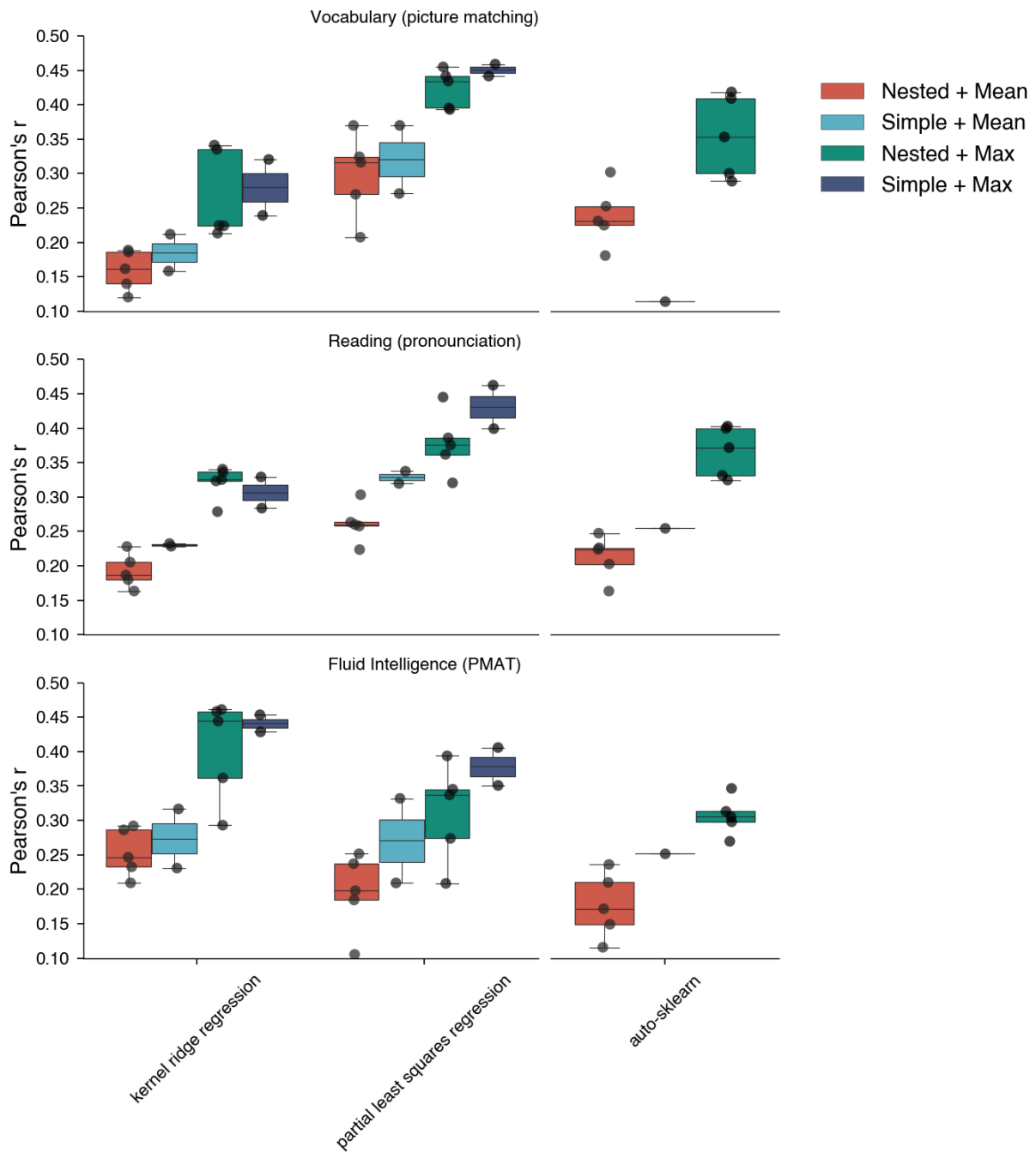

**Figure 6:** Effect of erroneous usage of CV for estimating generalization performance and reporting the maximum value across folds rather than the mean. Three targets (behavioral scores) were predicted in a 5-fold CV with 5 repeats using a subset of the HCP-YA S1200 release consisting of 369 unrelated subjects (192 males, 177 females) with ages ranging from 22 to 37 (*M*=28.63, SD=3.83). In one case the CV was used for hyperparameter tuning and estimation of generalization performance simultaneously (i.e., "simple" CV) and the mean values (light blue) and maximum values (dark blue) across folds were reported. In the other case nested CV was performed such that hyperparameter tuning was performed for each test fold independently in an inner CV loop applied on the training folds of the outer CV loop. Again, mean values and maximum values across folds are reported.

*3.b Test-to-test leakage*

Test-to-test leakage is a covert type of leakage that arises due to erroneous information sharing between the test samples. For the majority of ML applications samples in the test set should be treated independently. That is, processing and prediction of a given test sample should not depend on information from other test samples. For example, test-to-test leakage happens when the test set is used for estimating preprocessing parameters (Figure 7). In other words, instead of a single preprocessing model derived from the training samples (see Figure 1 for correct implementation) two preprocessing models are estimated: one using the training samples and applied to training samples, and another using the test samples and applied to test samples. For example, a practitioner may wish to demean their features. They may then estimate the mean of each feature in the training data and use these estimations to demean the training set. The error occurs if instead of also applying these estimations to the test data, the practitioner then estimates the mean values on the test data to demean the test data. This is wrong, because it implies that individual samples in the test data are demeaned depending on other samples in the test data. Another complication that arises in a pipeline that does not treat an individual test sample independently of other test samples, is that it cannot be deployed, and it might fail completely when applied to a single test sample. In the previous example of demeaning for instance, the demeaning could not be applied since the demeaned feature will be zero. In addition, these kinds of preprocessing steps attempt to estimate the parameters for a population, and estimates of parameters such as the mean are unlikely to be accurate from the typically smaller test set even if they did not result in impermissible operations.

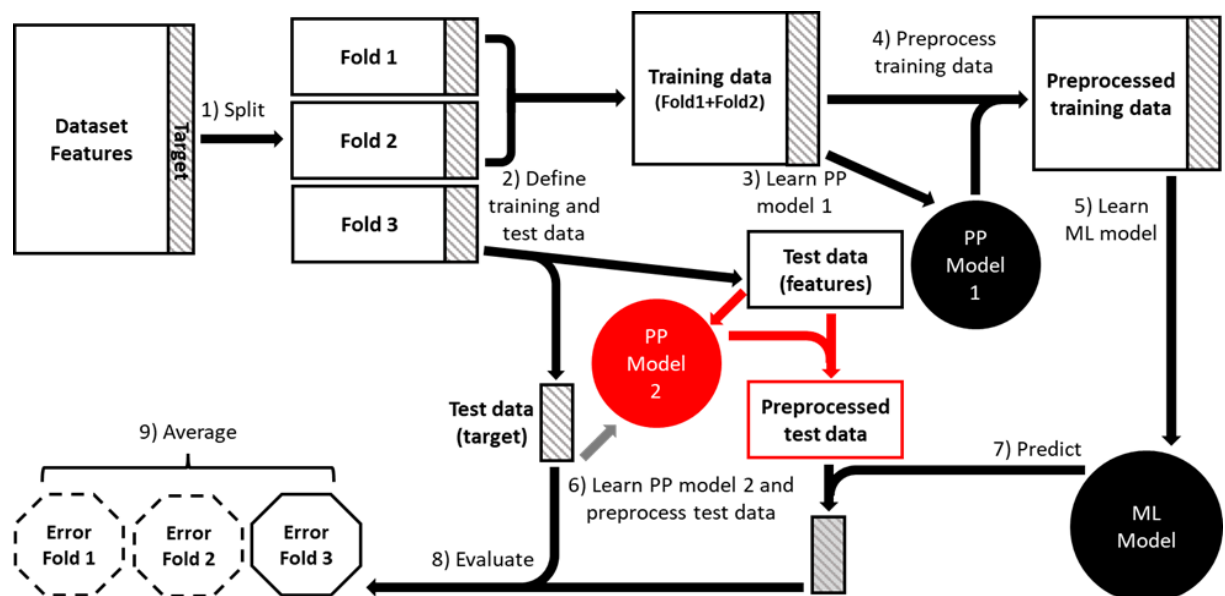

**Figure 7**: Test-to-test leakage. Preprocessing is performed in the test and training sets independently. Red color indicates the problematic steps.

Another case of test-to-test leakage can happen due to inappropriate use of the predictions obtained in CV. With repeated CV, which is usually recommended to avoid biases due to random data splitting, one obtains multiple predictions for each sample. Combining each sample's predictions across CV repeats, e.g., by averaging, causes leakage (Figure 8). Although this may seem like a rather elegant way to obtain a single prediction per sample and in turn a single error estimate, this procedure has two negative consequences. First, it generates ensemble results, and the performance reflects an ensemble of models built across repeats and not a single model as the practitioner intended and might claim. Second, it increases the effective number of CV folds ($k$) because across CV repeats the same data point is predicted using different combinations of training samples, effectively reducing the out-of-sample data points. With a high number of repetitions, the averaged prediction would be influenced by all other data points akin to leave-one-out CV, and the claim that the results reflect $k$-fold strategy would be wrong. These two reasons can lead to overoptimistic results.

We note that LOO per se is not problematic if properly implemented and with a correct evaluation metric that can be calculated for each sample separately and does not combine the samples. For instance, classification accuracy is suitable in a classification task, whereas AUROC is not. Similarly, when using LOO mean absolute error is appropriate but Pearson correlation should not be used.

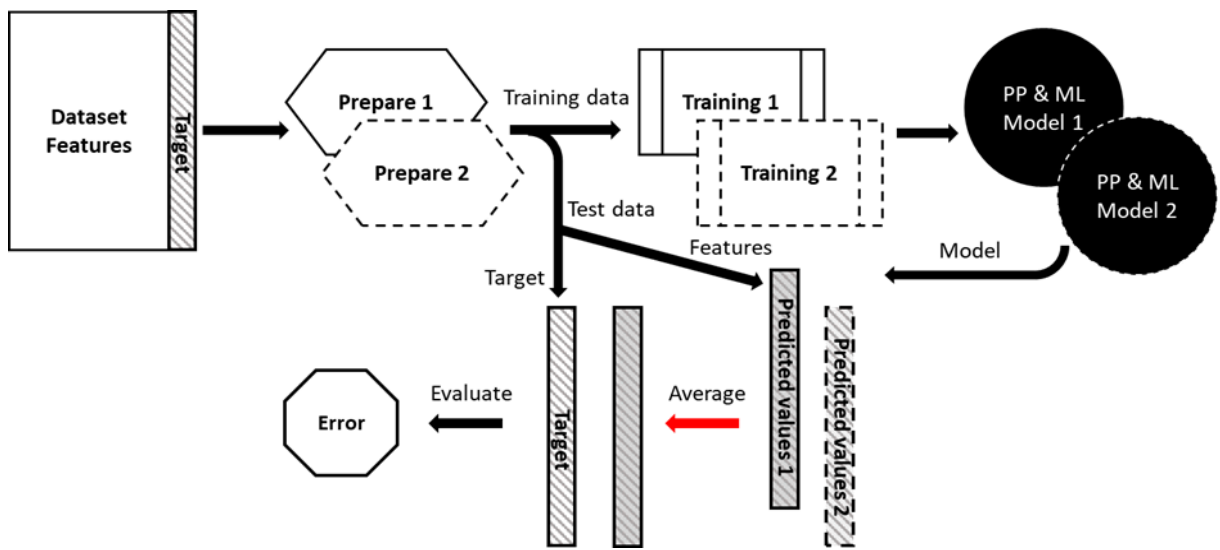

**Figure 8**: Test-to-test leakage due to averaging of predictions. For simplicity we show two runs of CV, and the major parts of CV are abstracted. Red color indicates the problematic steps.

*3.c Feature-to-target leakage*

Feature-to-target leakage occurs when the target is constructed using one or more features in the first place (Figure 9). For example, data is first clustered using some or all the features and the resulting cluster IDs are used as the target. Subsequent generalization estimation using CV on this data is likely optimistic as the supervised classification algorithm will simply need to reverse engineer the clustering process which it should be able to in most cases. Note that it is valid to use cluster ids as target to train a classification model and apply it to new unseen data.

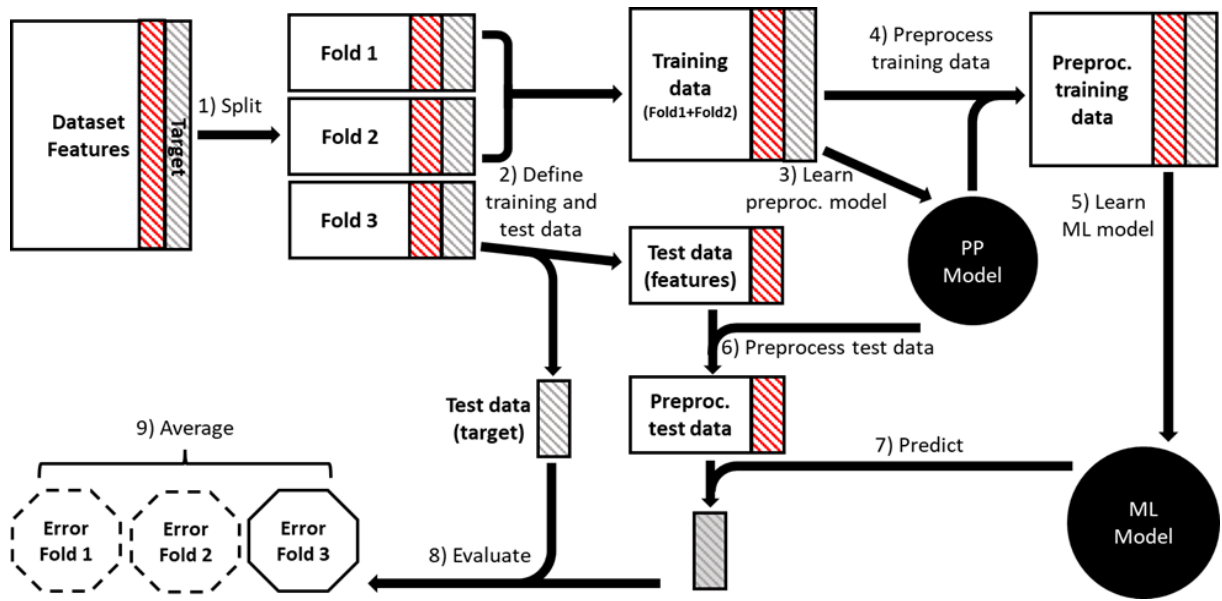

**Figure 9**: Feature-to-target leakage. The features include a variable, indicated with red color, that directly contributes to the target.

Feature-to-target leakage is also evident when one or more variables informative of the target are available during learning but not at the deployment phase. Since these features carry information that the models can learn from, it will lead to a high CV accuracy and the model might seem acceptable. However, this model is not deployable as the crucial features will not be available at the test time. As an example, consider a healthcare application where the goal is to diagnose a disease based on radiology data. To generate training data, radiology experts manually label each image as diseased or healthy. While labeling the data, the experts also make notes such as the size and location of the abnormality. If these notes are used together with the radiology images when building and validating models, it can lead to above mentioned issues. Given that these notes are predictive, the models will utilize them, and as a result, the CV will likely demonstrate high performance. However, these manually created notes will not be available during real-world use as the goal of such an application is to avoid manual annotation. Therefore, a model trained in this way cannot be deployed.

More generally, this type of leakage occurs due to differing train and the real test data distributions which again violates the I.I.D. assumption. However as CV is performed only in the training data the predictive information available is used providing an optimistic estimate of generalization performance. As such, this scenario is similar to reliance on confounds [2] and shortcut learning [12] where a model learns unintended signals prominently present in the training data which then hampers generalization to unseen data in which these signals are missing. While this scenario could be viewed as learning incorrect information rather than leakage, we categorize it as leakage because it results in overestimation of the model's ability to generalize which is typically the culmination of academic exercises. Note that feature-to-target leakage has also been termed as target leakage[2] but we reserve this term for another type of leakage described below.

*3.d Target leakage*

We define target leakage as the scenario where the target is utilized in the prediction process. This is an overt kind of leakage that can occur due to use of specific algorithms and preprocessing steps that rely on the target.

An example is misapplication of the partial least squares (PLS) algorithm, which is popular in several fields. PLS is a bilinear factor model that identifies a new space representing the covariance between the features $X$ and target $Y$ spaces. As PLS estimates a shared latent space, it generates beta values for both $X$ and $Y$. If the beta values $Y$ are also used instead of only that of $X$ to make predictions then one requires the target values of the test examples, resulting in data leakage. As when performing CV the target values of the test data are available one could provide them but this will result in leakage in addition to creating a non-deployable model. Thus when using PLS the practitioner should make sure that only $X$ values are used for prediction. Note that most libraries provide a correct implementation of PLS.

Another case of target-leakage can occur while trying to mitigate measurement bias by means of data harmonization. As data collection becomes more accessible and widespread, pooling data together from different sites has become increasingly common. Differences in data collection protocols and measurement devices can induce systematic biases. To address this, batch-effect removal [79] or data harmonization [31] is used. If the data are harmonized across sites (e.g., by matching covariances) too aggressively, it could also remove variance of interest that is related to the target. To avoid this, harmonization methods can preserve variance related to user-specified covariates which can also include the target. When harmonizing new data, the same covariates must be provided for each test sample. Effectively, while it is beneficial to preserve variance related to the target while harmonizing, the target values must be available when making predictions on test data. This presents two possible ways to implement harmonization while estimating generalization performance using CV, both of which lead to leakage. First, one can harmonize all the data before creating splits which violates the train-test data separation requirement. Second, the target values of the test set are needed for harmonization which amounts to target leakage and also precludes deployment. Overall, use of harmonization in a ML pipeline while explicitly preserving the target-related variance can be considered as target leakage.

*3.e Dataset leakage: Dataset decay*

In traditional statistics it is widely recognized that testing multiple hypotheses (e.g., mass-univariate hypothesis testing) can result in spurious discoveries, and hence approaches to correct for multiple comparisons are utilized. The multiple testing issue might not seem pertinent to an individual researcher who only aims to test a single hypothesis. On a larger scale, however, multiple researchers testing different hypotheses on the same dataset leads to an increased global likelihood of false positive findings, referred to as "dataset decay" [69]. We term this as dataset leakage as the information regarding the whole dataset is leaked across different analyses.

---

[2] http://downloads.alteryx.com/betawh_xnext/MachineLearning/MLTargetLeakage.htm
https://h2o.ai/wiki/target-leakage/

In some fields, for example neuroimaging, the availability of large datasets suitable for ML analysis is relatively limited, e.g. HCP and ADNI. Consequently, these datasets are extensively used by a multitude of researchers in the field. This widespread usage has sparked a competitive environment where there is a continual race to develop new methods that outperform existing state-of-the-art scores. However, this intense focus on a few datasets and high accuracy can lead to dataset decay such that the overuse and repeated analysis of the same datasets may diminish their effectiveness for future research and innovation.

ML studies testing multiple analysis pipelines on the same data lead to similar issues. Specifically, the best CV estimates from testing a large number of ML algorithms or pipelines on a given dataset are overly optimistic [51, 58]. While data sharing is beneficial for enhancing collaborative research and reproducibility, it extends dataset leakage to analyses performed by the community such that contemporary performance evaluations on re-used datasets lead to overfitting and overly optimistic estimates of generalizability [17, 24]. Furthermore, a researcher's data analysis strategy might be informed by previous analyses carried out on the same data. This not only increases the number of comparisons but also introduces dependency between analyses, frequently leading to further false discoveries and inflated effect sizes [27].

In fact, our empirical demonstrations may be a good example of such overfitting: In both Figure 4 and Figure 6, kernel ridge regression and partial least squares regression were used for predicting a number of behavioral scores based on functional connectivity. These two models are well known to work on this task and have been chosen for this exact reason [9, 29]. Importantly, they are known to work well on the exact dataset we have used here - the Human Connectome Project Young Adult. This could explain why instances of these models outperformed models found by Auto-ML. However, it is important to note that this conclusion is only suggestive and further evaluations using additional datasets are needed to assess whether the algorithms and resulting outcomes are truly overfitted.

3.f Confound leakage

A ML model can be influenced by confounding factors which in turn can influence its predictions. Confounding factors are related to both the features and the target. Depending on the goal of a study, it may or may not be desirable to minimize their impact. If the goal of a study is to simply predict a target as well as possible, then confounding may not necessarily be a worry [40]. But if researchers want to gain insight about the specific relationship between features and target independent of the confounding factors, strategies to mitigate the effect of confounders must be applied. A common example of a confounding factor in brain imaging studies is age. For instance, brain imaging can accurately reflect a person's age and can also contain information about age-related diseases [49]. The problem here is that the model simply learns how brain images may change with the natural aging process, but may not learn anything about the specific changes and processes related to pathology. From a prediction point of view this may be an issue in some cases, for instance a young person with pathology will not be identified as a patient. Furthermore, such a model is likely less helpful in gaining specific biological insight than a model that learns about processes specifically involved in the disease.

Thus, if not properly controlled or accounted for, confounding can lead to correct predictions, but may mislead the researcher to incorrectly conclude that their model is learning about the specific disease-related processes. For example, a particular feature of brain structure might be erroneously deemed important in the prediction of Parkinson's disease, whereas the association is actually due to the confounding effect of age. In other words, the brain structural feature changes with increasing age, and increasing age also leads to a higher likelihood of developing Parkinson's disease, but the structural feature in reality has nothing to do with the disease.

However, even if ML practitioners do not care about interpretability or insight with respect to their model and how it represents the relationship between features and target, confounding factors can be a problem. That is, confounding can also degrade the model's performance, especially when the model is deployed in environments where the distribution of the confounding variable is different. This is akin to the assumption that the data used to train a model should follow the same distribution as the data for which the model is deployed, with the only difference being that confounds in this scenario are not modeled explicitly (but importantly change the relationship between features and target). Thus, the model trained and evaluated in a particular distribution (of the confounding factor) will lead to overly optimistic estimates of the generalization error, that may not tell us about how the model will perform on data for which the distribution of the confounding factor is changed.

Feature-wise confound removal by means of linear regression (i.e. confound regression) is a standard method used to deal with confounding effects in retrospective data analyses which is a common scenario in ML [56, 64]. It is recommended to perform confound regression in a CV-consistent manner to avoid test-to-train data leakage [50]. However, the process of confound regression itself can leak information into the features especially when the feature-target distribution is skewed [26]. In this case, variance associated with the confounds is injected into the features rather than being removed as expected. This type of leakage becomes particularly problematic when the confounds are strongly associated with the target and the leakage increases with the number of features. Confound-leakage can lead to above-chance generalization performance estimates, even when the relationship between the features and the target is destroyed. This provides a method to check for potential confound leakage by shuffling the features and performing CV. If the CV accuracy is above-chance then one can conclude that confound leakage is possible on this dataset.

## 4. Possible mitigation strategies

In this section we provide advice to improve common reporting practices to facilitate the detection of leakage as well as to increase reproducibility in ML pipelines. First we should mention that excellent recommendations exist for reporting ML models such as Model Cards [48] and DOME [77]. Further guidelines have been proposed for specialized domains such as biomedical applications [46] and minimum information about clinical artificial intelligence modeling [52]. Data processing and quality reporting have also been discussed (e.g. Datasheets for Datasets [20] and Data Nutrition Labels [30]).

However, as the application contexts and modeling intricacies of ML-based analyses expand and grow in complexity, we think it is necessary to refine current recommendations, especially by making data processing and model selection and assessment strategies more transparent. Effective communication of ML pipelines can take various forms. Textual descriptions offer high-level overviews of a pipeline and its components, providing a conceptual understanding. However, such descriptions can fall short.

When documenting the experimental setup, it is crucial to provide comprehensive details. However, ambiguous descriptions hinder understanding and replicability. For instance, stating that "default (hyper)parameters" were utilized for a given algorithm is inadequate as default hyperparameters are not universally defined and can vary between software implementations and labs. Moreover, defaults may change over time and with software updates. Therefore, to guarantee replicability, it is essential to report the specifics of the set up including data processing, model hyperparameters, training and evaluation procedures, as well as the software packages used and their corresponding versions.

In addition, there are several instances where the method description, whether written or verbal, is insufficient to detect leakage. In some cases, the description seems to be correct, even when leakage is present. Therefore, researchers and reviewers should not only rely on the written description, but instead insist on reviewing the code written to implement the methods. To illustrate, let's look at this example method description: "We performed feature selection using LASSO followed by an SVM classifier with an RBF kernel. Both feature selection and hyperparameter tuning were performed on the training folds within a cross-validation loop". Based on this, a reader would assume that the procedures were correctly implemented. However, when we were unable to replicate the results on the same data, we decided to examine the code more closely. The issue became immediately apparent: test-to-train leakage during feature selection using LASSO. Although both steps, feature selection and hyperparameter optimization, were performed within a CV loop as mentioned in the text, these two steps were implemented in two separate CV loops (see Fig. 5). The first CV loop performed feature selection, correctly only on the training folds, and noted which features were selected (nonzero LASSO weights). The features selected more than once across CV folds were retained. Then in another CV loop, an SVM classifier was trained on the training folds with hyperparameter tuning while using only the selected features and tested on the test fold. It is clear that this procedure causes leakage even though the text description can be interpreted as being correct. It is crucial to note that the feature selection method (as long as it is data driven) or the fact that the same CV fold structure was used for the two loops, does not prevent the leakage. What is problematic is the first CV loop where choice of which features to use was made using all the data. Therefore, the textual description of the process was insufficient to detect the issue of leakage. However, a quick examination of the underlying code clarified the situation. This highlights the importance of transparency in machine learning research and practice, and we strongly encourage researchers and practitioners to share their code. In addition, to avoid such errors before publication it is crucial that labs implement their own standard procedures to perform adequate code review. Since internal code review might not catch all errors, a greater emphasis should be put by journals and reviewers on reviewing code during the peer-review process.

Sharing source code allows for a detailed examination of the implementation, enhancing transparency and enabling replication or modification. While sharing their code may indeed expose it to critical review, it is essential for progress in this predominantly software-driven discipline [3, 63]. There may be errors spotted and improvements suggested, but this iterative process of refinement is an integral part of scientific advancement.

In this regard, we echo the call to make research code openly available. This practice would aptly put importance on correctness of the pipelines and ensuing analyses. The authors admit that they themselves have done this with less stringency in the past, but we aim to do better. We also recommend that ML practitioners, especially in early career stages, request code reviews to identify and fix any issues with their implementation. Importantly, however, the responsibility should not be off-loaded to early career researchers and standard procedures (to ensure good quality code and research) need to be implemented at an organizational level. For example, each lab can perform code reviews using one or more skilled reviewers. Such a code review should focus on the correctness of the code itself and not on its functionality or whether it produces the desirable output such as high accuracy. However, we recognize that it can be challenging to find skilled individuals that can perform a proper review.

Openly sharing data, trained models, and other research artifacts/tools fosters reproducibility, and encourages collaboration, enabling the broader community to validate and build upon the work. However, it must be noted that the shared data should not be affected by data leakage (e.g., during preprocessing). Detecting such instances of leakage can be challenging, if not impossible once the data is shared.

## 5. Conclusions

Data leakage presents a significant challenge in machine learning. Identifying and preventing leakage is essential for ensuring reliable and robust models. To this end, we have provided several examples and detailed explanations of data leakage instances, along with tips on how to identify them.

Below, we present a few crucial points that underlie most of the leakage cases we have presented in this article.

- Ensure strict training-test set separation.

- Ensure that performance metrics are calculated on truly unseen data that has not been used anywhere in the pipeline previously.

- Model selection and model assessment should be done with a nested CV.

- State the goal of your ML pipeline: Search for a feature-target relationship, assess generalization performance or deployment? Clarifying the goals early on will help practitioners to design and implement a correct and appropriate pipeline.

- While academic applications of machine learning often do not involve deploying models, considering whether a pipeline can actually be used to make predictions on genuinely unseen data can aid in identifying potential instances of data leakage. Of note, check if features are available after deployment. Can a model be applied to future test data not currently available? Can it be applied to a single test example?

- Detailed description of methods as well as sharing your code publicly is an effective way to ensure transparency in your pipeline design. In addition, releasing your models enables users to test data gathered after the model's publication, which further boosts confidence in the model's ability to generalize well.

- When possible, opt for using well-established software packages and libraries instead of creating standard procedures from scratch. We certainly do not discourage code implementation for learning purposes, but we recognize that code testing can be time-consuming and challenging. Hence, using standardized code in production environments is typically a more effective choice.

- Ensuring the correctness of a ML pipeline should take precedence over its output or even its replicability. A flawed pipeline might yield accurate and replicable results, but that should not be mistaken as an indication of the validity of the models or the ensuing results.

Finally, in addition to leakage, several other pitfalls and issues exist and deserve attention, (real-world usefulness of benchmark data [65], dataset biases [22, 44] and deployment challenges [53]). However, it is not possible to cover all those aspects in a single paper. We recommend that readers stay vigilant and pay attention to issues that might affect their specific analysis set up.

## 6. Acknowledgements

This research was partially funded by the Deutsche Forschungsgemeinschaft (DFG, German Research Foundation) – Project-ID 431549029 - Collaborative Research Centre CRC1451 on motor performance project B05, by the Helmholtz Portfolio Theme "Supercomputing and Modelling for the Human" and the European Union's Horizon 2020 Research and Innovation Programme under Grant Agreement No. 945539 (HBP SGA3) and by the Max Planck School of Cognition supported by the Federal Ministry of Education and Research (BMBF) and the Max Planck Society (MPG).